%% file: main.tex
\PassOptionsToPackage{table}{xcolor}
\documentclass[10pt,twocolumn,letterpaper]{article}

\usepackage[pagenumbers]{cvpr} 

\input{preamble}
\usepackage{stfloats}
\definecolor{cvprblue}{rgb}{0.21,0.49,0.74}
\usepackage[pagebackref,breaklinks,colorlinks,allcolors=cvprblue]{hyperref}

\def\confName{CVPR}
\def\confYear{2027}
\newcommand{\ours}{DINO-KNN}
\newcommand{\nref}{R}  
\newcommand{\pmt}{\ensuremath{\pm}}
\newcommand{\R}{\mathbb{R}}
\usepackage{xcolor}

\definecolor{oursrow}{gray}{0.92}

\title{DINO-KNN: Training-Free Multiclass In-Context Segmentation\\via Nearest Neighbors}

\author{Hussain Alasmawi \quad Numan Saeed \quad Mohammad Yaqub\\
Division of Computing and Mathematical Sciences, Mohamed bin Zayed University\\
of Artificial Intelligence (MBZUAI), Abu Dhabi, United Arab Emirates
}

\begin{document}
\maketitle
\input{figures/teaser}
\input{sec/0_abstract}
\input{sec/1_intro}
\input{sec/2_related}
\input{sec/3_method}
\input{sec/4_experiments}

\input{sec/5_discussion}
\input{sec/6_conclusion}

\input{sec/7_llm}

{
    \small
    \bibliographystyle{ieeenat_fullname}
    \bibliography{ref.bib}
}

\clearpage
\setcounter{section}{0}\setcounter{table}{0}\setcounter{figure}{0}
\renewcommand{\thesection}{\Alph{section}}
\renewcommand{\thetable}{S\arabic{table}}
\renewcommand{\thefigure}{S\arabic{figure}}
\renewcommand{\pmt}[1]{{\tiny$\pm$#1}}   
\maketitlesupplementary
\input{sec/X_suppl}
\end{document}

%% file: preamble.tex
\usepackage{graphicx}
\usepackage{amsmath}
\usepackage{amssymb}
\usepackage{booktabs}

\usepackage{multirow}
\usepackage{makecell}
\usepackage{array}
\usepackage{subcaption}
\usepackage{cuted}

%% file: figures/teaser.tex
\begin{strip}
  \centering
  \vspace{-1.2em}
  \includegraphics[width=0.78\linewidth]{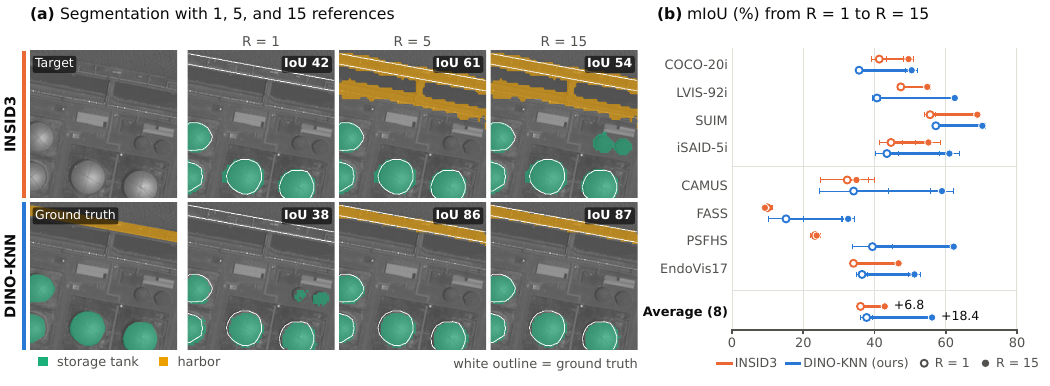}
  \captionsetup{hypcap=false}
  \captionof{figure}{\textbf{The accuracy of \ours{} grows with the number of reference ($\nref$) images.}
(a) A target segmented with $\nref{=}1$, $5$, and $15$ reference images.
(b) mIoU at $\nref{=}1$ and $\nref{=}15$ on eight datasets over three reference seeds.
On average, the mIoU rises by $18.4$ points for \ours{} and by $6.8$ for INSID3.}
  \captionsetup{hypcap=true}
  \label{fig:teaser}
\end{strip}

%% file: sec/0_abstract.tex
\begin{abstract}
In-context segmentation (ICS) segments a target image given a few annotated reference images. Recent training-free methods built on frozen foundation models segment one class at a time and gain little from additional references. We study the multiclass, multi-reference setting and ask how far a simple joint decision rule can go. We introduce \textbf{DINO-KNN}, which stores every labeled reference patch of a frozen DINOv3 encoder in a single memory and labels each target patch by a distance-weighted vote of its $k$ nearest neighbors. All classes are predicted in a single pass, with one label per pixel, without training or a mask decoder. We evaluate on eight datasets, four natural and four medical, with $\nref{=}1$ to 15 reference images and three reference seeds. DINO-KNN achieves the best average mIoU at every $\nref$; it is the best method on six of the eight datasets from $\nref{=}3$ and on seven of the eight from $\nref{=}7$. Its accuracy continues to grow with the number of references, from 37.8 mIoU at $\nref{=}1$ to 56.2 at $\nref{=}15$, while the strongest baseline, INSID3, rises from 36.1 to 42.9. DINO-KNN is also $2.1\times$ faster than INSID3. The code will be made public.
\end{abstract}

%% file: sec/1_intro.tex
\section{Introduction}
\label{sec:introduction}

Semantic segmentation assigns a semantic label to every pixel of an image.
Supervised models are trained to predict a predefined set of
classes~\cite{fcn,cheng2022masked,ronneberger2015u}, and dense annotations are
costly to obtain at scale~\cite{cityscapes}, especially in medical imaging, where they require expert annotators~\cite{tajbakhsh2020embracing}. In-context segmentation (ICS)
specifies the task through annotated examples instead: given a few reference
images with masks at inference time, a model segments the corresponding
concepts in a target image~\cite{seggpt,segic}. The reference annotations
define the task, which may involve one or several foreground classes.

Early ICS methods learn the mapping from reference annotations to target masks.
SegIC~\cite{segic} trains a mask decoder on top of a frozen encoder, and
SegGPT~\cite{seggpt} trains a generalist model on diverse segmentation
datasets, with a color-based formulation that supports several classes in one
reference. Neither updates its parameters at inference, but both require
supervised segmentation training.

Training-free methods remove this requirement by composing frozen foundation
models. Matcher~\cite{liu2024matcher} and GF-SAM~\cite{zhang2024gfsam}
use DINOv2~\cite{oquab2024dinov2} features to establish reference--target
correspondences, turn them into prompts, and obtain masks from the Segment
Anything Model (SAM)~\cite{kirillov2023sam}. Running two foundation models adds inference
cost, and prompt generation and mask selection add processing stages.
More recently, INSID3~\cite{cuttano2026insid3} showed that frozen
DINOv3~\cite{simeoni2025dinov3} features alone support training-free ICS,
without a learned decoder or a second foundation model. Its decision still
passes through several stages: fine-grained clustering, candidate localization
through backward correspondences, seed-cluster selection, and cluster
aggregation.

These training-free methods share a binary formulation: they predict one
foreground class at a time. Applied to a multiclass target, they produce one
mask per class, and these masks may assign conflicting labels to the same
pixel, so an additional rule is needed to obtain
mutually exclusive predictions. They also summarize the references of a class
in a compact form, such as a prototype or an averaged similarity map, and in our experiments they gain little beyond five references (\cref{tab:kscaling}).

This motivates our central question: \emph{can direct non-parametric
classification of frozen DINOv3 features provide accurate, training-free
multiclass ICS, without cluster aggregation or a learned segmentation decoder?}
We study this question in the multiclass, multi-reference setting, with several foreground classes per episode and one to fifteen reference images.

We introduce \textbf{DINO-KNN}, a training-free method for joint multiclass
ICS. Given annotated reference images, DINO-KNN stores every
labeled reference patch feature in a single memory and labels each target
patch by a distance-weighted vote of its \(k\) nearest neighbors (\(k\)-NN).
Keeping the individual reference features preserves the within-class
appearance variation captured by the encoder, and voting over several
neighbors reduces the sensitivity to a single wrong correspondence. All
classes compete in one decision that assigns each target patch one label, so
the predictions are mutually exclusive by construction and no binary masks
have to be merged. We adopt the positional debiasing of
INSID3~\cite{cuttano2026insid3}; the method has no other component.
Because every reference patch is kept, the accuracy of \ours{} keeps growing as references are added (\cref{fig:teaser}).
Our contributions are summarized as follows:

\begin{itemize}
\item We introduce \textbf{DINO-KNN}, a training-free and decoder-free method
for joint multiclass ICS based on direct \(k\)-NN classification of frozen
DINOv3 patch features. All classes are predicted in one pass with one label
per pixel.

\item We evaluate on eight datasets, with one to fifteen reference images. DINO-KNN achieves the best mIoU across the eight datasets at every \(\nref\). Its margin over the
strongest baseline, INSID3, grows from \(1.7\) mIoU with one reference to \(13.3\) with fifteen. Ablations show that
keeping individual reference patches, rather than a class prototype, lets \ours{} benefit from more references.

\item We show that the multi-stage pipelines of existing training-free methods are unnecessary for multiclass ICS: \ours{} labels all classes in a single \(k\)-NN step, is \(2.1\times\) faster than INSID3, and remains more accurate than INSID3 with smaller encoders.
\end{itemize}

%% file: sec/2_related.tex
\section{Related Work}
\label{sec:related_work}

\begin{figure*}[t]
  \centering
  \includegraphics[width=\linewidth]{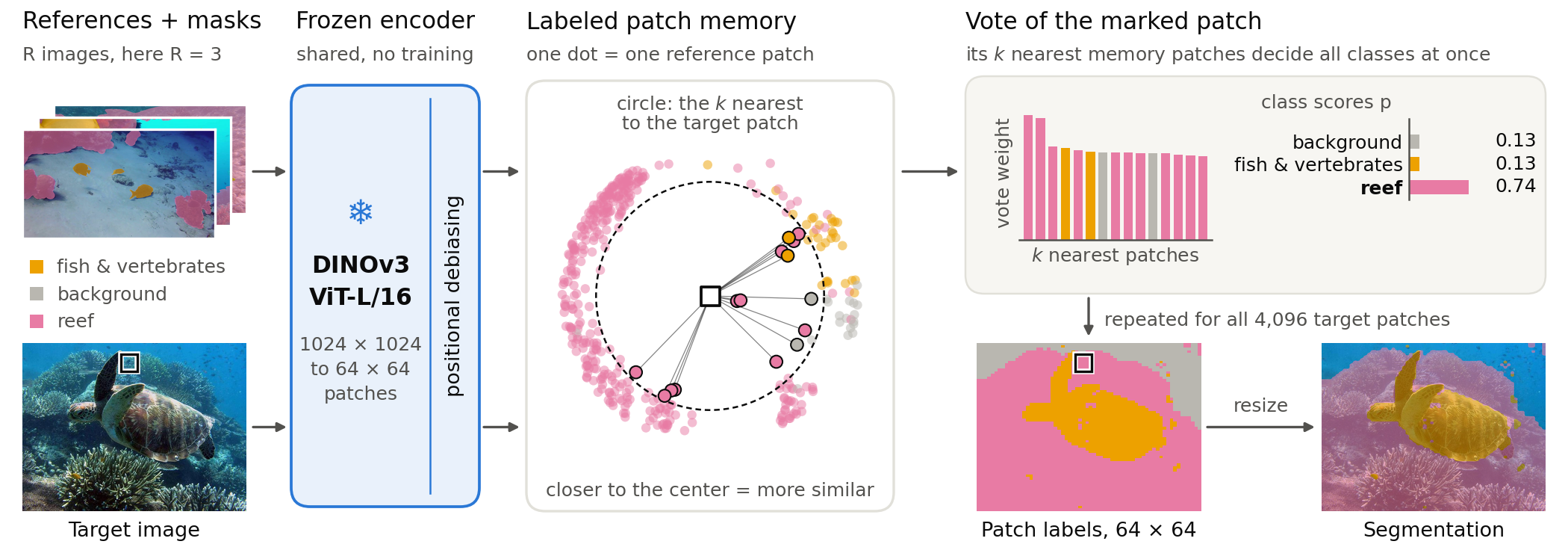}
  \caption{\textbf{Overview of \ours{}} with $\nref{=}3$ references.
  References and target are encoded by a frozen DINOv3 ViT-L/16 and positionally debiased.
  All labeled reference patches form one memory.
  The memory panel shows the reference patches around the marked target patch: a dot closer to the center is more similar to it, and the circle encloses its $k$ nearest patches.
  Their inverse-distance votes yield the patch's class scores.
  The vote is repeated for every target patch, and the patch label map is resized to the image.}
  \label{fig:framework}
\end{figure*}

\paragraph{Learned in-context and few-shot segmentation.}
Few-shot segmentation trains a network on base classes and tests it on novel classes given a few support images~\cite{shaban2017oslsm,panet,min2021hsnet}.
PANet~\cite{panet} represents each class by a prototype, the masked average of the support features, in the spirit of prototypical networks~\cite{snell2017prototypical}.
In-context models extend this idea to large-scale training.
Visual prompting by inpainting~\cite{bar2022visual}, Painter~\cite{Wang_2023_CVPR} and SegGPT~\cite{seggpt} cast segmentation as image completion; SegIC~\cite{segic} and SINE~\cite{liu2024sine} train lightweight decoders on top of frozen backbones; UniverSeg~\cite{butoi2023universeg} and SSL-ALPNet~\cite{ouyang2020self} target medical images.
All of these need a training stage on annotated segmentation data.
We restrict our study to methods that require no training.

\paragraph{Training-free in-context segmentation.}
PerSAM~\cite{zhang2024persam} locates the target object with one reference feature and prompts SAM~\cite{kirillov2023sam}.
Matcher~\cite{liu2024matcher} matches reference and target patches with DINOv2 features~\cite{oquab2024dinov2}, keeps only the consistent matches as point prompts, and selects among many SAM mask proposals.
GF-SAM~\cite{zhang2024gfsam} replaces the proposal search with a graph over point prompts and masks.
INSID3~\cite{cuttano2026insid3} shows that DINOv3 features~\cite{simeoni2025dinov3} alone suffice: it removes a positional component from the features, clusters the target, and selects and aggregates clusters by their similarity to a reference prototype.
All four methods segment in several stages: feature matching, prompt or cluster generation, and mask selection or aggregation, each with its own design choices.
They produce one binary mask per class and are evaluated with one or a few references.
\ours{} replaces these stages with a single nearest-neighbor vote that labels all classes at once, and we study it together with the four methods on a larger number of references.

\paragraph{Non-parametric prediction on frozen features.}
$k$-NN classification is a standard probe of self-supervised features~\cite{wu2018instance,caron2021dino,oquab2024dinov2}.
Copying labels from the nearest patches of annotated frames is also the standard protocol for evaluating features on video object segmentation~\cite{jabri2020walk,caron2021dino}.
Hummingbird~\cite{balazevic2023hummingbird} applies nearest-neighbor retrieval to dense prediction: patch labels are propagated from a memory built from a training set, with an encoder pre-trained for this use.
Our decision rule belongs to this family, but the setting differs: the memory holds only the patches from the $\nref$ reference images for an episode rather than a training set, and the label set changes from one episode to the next.

%% file: sec/3_method.tex
\section{Method}
\label{sec:method}

Given a few annotated reference images and an unlabeled target image, our goal is to segment the target into the classes specified by the reference annotations.
We introduce \ours{} (\cref{fig:framework}), a training-free and non-parametric method that performs this task with frozen DINOv3 patch features and $k$-nearest-neighbor ($k$-NN) classification.
\ours{} extracts patch features from all images and applies the positional debiasing of INSID3~\cite{cuttano2026insid3} (\cref{sec:features}).
It then pairs each reference patch feature with its label and collects these pairs into a single memory (\cref{sec:memory}).
For each target patch, \ours{} computes class scores by a $k$-NN vote over this memory and assigns the class with the highest score (\cref{sec:decision}).
The result is one multiclass label map, obtained without a learned segmentation decoder.

\subsection{Problem setting}
\label{sec:setting}
An episode consists of a target image $I_t$, a set of $\nref$ reference images with masks $\{(I_r, Y_r)\}_{r=1}^{\nref}$, and a label set $\mathcal{C}=\{1,\dots,C\}$ of classes.
Every pixel of a reference mask carries one label of $\mathcal{C}$.
The task is to predict one label per target pixel,
\begin{equation}
  \hat{Y}_t = f\big(I_t \mid \{(I_r, Y_r)\}_{r=1}^{\nref}\big) \in \mathcal{C}^{H\times W},
\end{equation}
where $H{\times}W$ is the target size.
The function $f$ has no trainable parameters and sees no data other than the episode.

\subsection{Patch features and positional debiasing}
\label{sec:features}
Every image is resized to a fixed resolution and encoded by a frozen DINOv3 encoder~\cite{simeoni2025dinov3}.
We take the patch tokens of the last block and $\ell_2$-normalize them,
\begin{equation}
  F = \Phi(I) \in \R^{P\times D},
\end{equation}
where $P$ is the number of patches and $D$ the feature dimension.

DINOv3 features include a positional component that allows patches at similar locations to match even when their content differs~\cite{cuttano2026insid3}.
We remove it using the INSID3 debiasing procedure.
Let $F_0=\Phi(I_0)$ be the features of a Gaussian noise image $I_0$ and $F_0 = U\Sigma V^{\top}$ their singular value decomposition.
The first $d_{\text{pos}}$ right singular vectors $B = V_{[:,1:d_{\text{pos}}]}\in\R^{D\times d_{\text{pos}}}$ span the estimated positional subspace.
Every feature matrix is projected onto the orthogonal complement of this subspace and normalized again:
\begin{equation}
  \tilde{F} = \operatorname{norm}\big(F\,(I_D - BB^{\top})\big).
\end{equation}
Here, $I_D$ is the $D{\times}D$ identity matrix, and $\tilde{F}$ contains the debiased patch features.
The basis $B$ is computed once and reused for all images and datasets.

\subsection{Labeled patch memory}
\label{sec:memory}
Each reference mask is resized to the patch grid with nearest-neighbor interpolation.
The memory is the set of all labeled reference patches of the episode,
\begin{equation}
  \mathcal{M} = \big\{(\tilde{f}_i, y_i)\big\}_{i=1}^{N}, \qquad y_i\in\mathcal{C}, \quad N = \nref P .
\end{equation}
Here $\mathcal{M}$ is the memory, $N$ is the number of entries, and $\tilde{f}_i$ and $y_i$ are the debiased feature and the label of the $i$-th reference patch.
The patches of a class are not averaged into a prototype; each one is kept as a separate entry.

\begin{table*}[t]
  \centering
  \caption{\textbf{Multiclass mIoU for $\nref{=}1$ to $15$ reference images}, mean {\tiny$\pm$std} over three seeds.
  PerSAM accepts a single reference.
  Best mean per column and $\nref$ in bold.}
  \label{tab:kscaling}
  \scriptsize
  \setlength{\tabcolsep}{2.6pt}
  \renewcommand{\arraystretch}{0.93}
  \input{tab/kscaling}
\end{table*}

\subsection{Nearest-neighbor decision}
\label{sec:decision}
Each target patch receives the label that dominates among the reference patches most similar to it.
The decision has three steps.

\paragraph{Search.}
For a target patch $j$ with feature $\tilde{f}_j$, we compute the cosine similarity $s_{ji}=\tilde{f}_j^{\top}\tilde{f}_i$ to every memory entry $i$ and keep the $k$ most similar entries, $\mathcal{N}_k(j)$.
The search runs over the entire memory, so the neighbors can belong to any class.

\paragraph{Vote.}
Each neighbor votes for its own label, and a more similar neighbor carries more weight.
The weight of a vote is the inverse of the cosine distance~\cite{dudani1976weighted}, and the score of class $c$ is its share of the total weight:
\begin{equation}
  w_{ji} = \frac{1}{1-s_{ji}}, \quad
  p_j(c) = \frac{\sum_{i\in\mathcal{N}_k(j)} w_{ji}\,\mathbf{1}[y_i=c]}{\sum_{i\in\mathcal{N}_k(j)} w_{ji}} .
\end{equation}

\paragraph{Label.}
The patch takes the class with the highest score,
\begin{equation}
  \hat{y}_j = \arg\max_{c\in\mathcal{C}}\, p_j(c),
  \label{eq:decision}
\end{equation}
and the resulting label map is resized to the original image size with nearest-neighbor interpolation.

\paragraph{Properties.}
A single vote decides all classes, so every pixel receives exactly one label.
No class is processed separately, and no overlapping masks have to be merged.
The computation consists of a matrix product between the target patches and the memory, followed by a top-$k$ selection.
Its cost does not depend on the number of classes and grows linearly with $\nref$.

%% file: tab/kscaling.tex
\begin{tabular}{@{}cl cccc cccc ccc@{}}
\toprule
 & & \multicolumn{4}{c}{Natural images} & \multicolumn{4}{c}{Medical images} & \multicolumn{3}{c}{Average} \\
\cmidrule(lr){3-6}\cmidrule(lr){7-10}\cmidrule(l){11-13}
$\nref$ & Method & COCO & LVIS & SUIM & iSAID & CAMUS & FASS & PSFHS & EndoVis17 & Natural (4) & Medical (4) & All (8) \\
\midrule
\multirow{4}{*}{1} & PerSAM & 6.7\pmt{2.1} & 6.9\pmt{0.9} & 20.9\pmt{1.0} & 17.2\pmt{4.9} & 27.1\pmt{11.0} & 6.9\pmt{4.2} & 20.7\pmt{2.1} & 17.0\pmt{0.7} & 12.9\pmt{2.2} & 17.9\pmt{2.4} & 15.4\pmt{2.2} \\
 & GF-SAM & 41.2\pmt{1.2} & 43.5\pmt{1.0} & \textbf{58.9}\pmt{0.9} & 37.7\pmt{1.3} & 30.1\pmt{5.0} & 7.8\pmt{1.0} & 22.3\pmt{2.7} & 34.8\pmt{1.7} & 45.3\pmt{0.4} & 23.7\pmt{1.9} & 34.5\pmt{0.8} \\
 & INSID3 & \textbf{41.3}\pmt{2.1} & \textbf{47.4}\pmt{0.9} & 55.6\pmt{1.5} & \textbf{44.6}\pmt{3.3} & 32.3\pmt{7.5} & 9.9\pmt{1.4} & 23.4\pmt{1.5} & 34.1\pmt{0.3} & \textbf{47.2}\pmt{0.4} & 24.9\pmt{2.0} & 36.1\pmt{0.8} \\
\rowcolor{oursrow} & \textbf{DINO-KNN} (ours) & 35.7\pmt{0.8} & 40.7\pmt{1.1} & 57.2\pmt{0.4} & 43.5\pmt{3.3} & \textbf{34.1}\pmt{9.7} & \textbf{15.2}\pmt{4.9} & \textbf{39.4}\pmt{5.6} & \textbf{36.5}\pmt{1.5} & 44.3\pmt{1.0} & \textbf{31.3}\pmt{2.8} & \textbf{37.8}\pmt{1.7} \\
\midrule
\multirow{3}{*}{3} & GF-SAM & \textbf{48.5}\pmt{1.6} & 51.4\pmt{1.0} & \textbf{65.9}\pmt{1.0} & 43.8\pmt{2.1} & 32.2\pmt{4.5} & 9.3\pmt{3.5} & 21.9\pmt{0.6} & 38.4\pmt{0.5} & 52.4\pmt{0.8} & 25.4\pmt{0.7} & 38.9\pmt{0.1} \\
 & INSID3 & 46.5\pmt{2.5} & 52.6\pmt{0.8} & 63.8\pmt{0.4} & 51.3\pmt{3.1} & 33.0\pmt{4.7} & 8.4\pmt{2.7} & 22.6\pmt{0.9} & 39.9\pmt{0.4} & 53.6\pmt{0.1} & 26.0\pmt{1.3} & 39.8\pmt{0.6} \\
\rowcolor{oursrow} & \textbf{DINO-KNN} (ours) & 44.7\pmt{1.1} & \textbf{53.1}\pmt{1.2} & 65.5\pmt{0.5} & \textbf{53.7}\pmt{2.0} & \textbf{45.2}\pmt{10.3} & \textbf{16.4}\pmt{5.7} & \textbf{50.8}\pmt{2.6} & \textbf{43.8}\pmt{1.2} & \textbf{54.2}\pmt{0.8} & \textbf{39.0}\pmt{1.9} & \textbf{46.6}\pmt{0.8} \\
\midrule
\multirow{3}{*}{5} & GF-SAM & \textbf{49.6}\pmt{2.0} & 53.0\pmt{1.3} & \textbf{67.2}\pmt{0.7} & 45.4\pmt{1.9} & 32.9\pmt{2.4} & 10.7\pmt{0.6} & 22.6\pmt{0.9} & 39.4\pmt{0.8} & 53.8\pmt{1.0} & 26.4\pmt{0.9} & 40.1\pmt{0.1} \\
 & INSID3 & 48.2\pmt{1.5} & 54.1\pmt{0.8} & 66.3\pmt{0.6} & 53.3\pmt{3.5} & 34.4\pmt{3.7} & 10.7\pmt{0.7} & 23.3\pmt{1.6} & 42.2\pmt{2.0} & 55.5\pmt{0.8} & 27.6\pmt{0.7} & 41.6\pmt{0.2} \\
\rowcolor{oursrow} & \textbf{DINO-KNN} (ours) & 46.1\pmt{1.8} & \textbf{56.9}\pmt{0.5} & 67.1\pmt{0.7} & \textbf{56.4}\pmt{2.8} & \textbf{50.0}\pmt{3.5} & \textbf{24.5}\pmt{5.3} & \textbf{53.8}\pmt{3.5} & \textbf{46.4}\pmt{1.8} & \textbf{56.7}\pmt{0.7} & \textbf{43.7}\pmt{1.6} & \textbf{50.2}\pmt{1.1} \\
\midrule
\multirow{3}{*}{7} & GF-SAM & \textbf{50.5}\pmt{1.6} & 53.7\pmt{1.3} & 68.2\pmt{1.1} & 46.5\pmt{2.5} & 33.1\pmt{2.3} & 10.2\pmt{0.5} & 23.1\pmt{0.9} & 39.8\pmt{0.1} & 54.7\pmt{1.0} & 26.6\pmt{0.8} & 40.6\pmt{0.2} \\
 & INSID3 & 49.2\pmt{1.0} & 54.2\pmt{0.9} & 66.7\pmt{0.3} & 53.7\pmt{3.9} & 34.3\pmt{3.0} & 10.0\pmt{0.6} & 23.7\pmt{1.5} & 42.1\pmt{1.3} & 56.0\pmt{1.1} & 27.5\pmt{0.7} & 41.7\pmt{0.3} \\
\rowcolor{oursrow} & \textbf{DINO-KNN} (ours) & 48.2\pmt{1.0} & \textbf{59.1}\pmt{0.6} & \textbf{68.3}\pmt{0.2} & \textbf{56.4}\pmt{2.3} & \textbf{52.1}\pmt{4.0} & \textbf{26.4}\pmt{5.8} & \textbf{59.7}\pmt{2.0} & \textbf{48.5}\pmt{1.5} & \textbf{58.0}\pmt{0.8} & \textbf{46.7}\pmt{1.5} & \textbf{52.3}\pmt{0.8} \\
\midrule
\multirow{3}{*}{9} & GF-SAM & \textbf{50.9}\pmt{1.5} & 53.8\pmt{1.1} & 68.9\pmt{1.6} & 47.0\pmt{2.7} & 33.1\pmt{2.5} & 10.1\pmt{0.6} & 22.6\pmt{0.6} & 39.4\pmt{0.8} & 55.2\pmt{1.0} & 26.3\pmt{0.7} & 40.8\pmt{0.3} \\
 & INSID3 & 48.6\pmt{0.4} & 55.0\pmt{0.8} & 67.4\pmt{0.3} & 54.6\pmt{3.8} & 34.1\pmt{3.2} & 9.6\pmt{0.4} & 23.4\pmt{1.2} & 44.2\pmt{1.1} & 56.4\pmt{1.2} & 27.8\pmt{1.2} & 42.1\pmt{0.4} \\
\rowcolor{oursrow} & \textbf{DINO-KNN} (ours) & 49.4\pmt{1.3} & \textbf{60.4}\pmt{0.7} & \textbf{69.0}\pmt{0.7} & \textbf{58.0}\pmt{2.4} & \textbf{54.2}\pmt{4.9} & \textbf{30.1}\pmt{2.5} & \textbf{60.3}\pmt{1.4} & \textbf{49.2}\pmt{1.7} & \textbf{59.2}\pmt{0.9} & \textbf{48.4}\pmt{1.1} & \textbf{53.8}\pmt{0.2} \\
\midrule
\multirow{3}{*}{11} & GF-SAM & \textbf{51.6}\pmt{2.7} & 54.4\pmt{0.4} & 68.9\pmt{1.2} & 47.3\pmt{2.3} & 33.7\pmt{2.4} & 10.4\pmt{0.6} & 23.1\pmt{0.4} & 40.3\pmt{0.4} & 55.6\pmt{1.1} & 26.9\pmt{0.7} & 41.2\pmt{0.2} \\
 & INSID3 & 49.2\pmt{1.2} & 55.2\pmt{1.1} & 67.7\pmt{0.9} & 54.8\pmt{3.4} & 34.5\pmt{2.7} & 9.2\pmt{0.5} & 23.9\pmt{1.0} & 45.7\pmt{1.5} & 56.7\pmt{1.2} & 28.3\pmt{0.9} & 42.5\pmt{0.4} \\
\rowcolor{oursrow} & \textbf{DINO-KNN} (ours) & 49.4\pmt{1.8} & \textbf{61.1}\pmt{0.3} & \textbf{69.5}\pmt{0.9} & \textbf{59.4}\pmt{2.6} & \textbf{56.9}\pmt{5.5} & \textbf{32.0}\pmt{2.4} & \textbf{61.1}\pmt{1.3} & \textbf{49.8}\pmt{1.9} & \textbf{59.9}\pmt{0.9} & \textbf{50.0}\pmt{1.7} & \textbf{54.9}\pmt{0.7} \\
\midrule
\multirow{3}{*}{13} & GF-SAM & \textbf{51.7}\pmt{2.4} & 54.3\pmt{0.9} & 68.7\pmt{1.3} & 47.6\pmt{2.3} & 34.1\pmt{2.5} & 11.2\pmt{0.5} & 23.0\pmt{0.1} & 40.1\pmt{0.2} & 55.5\pmt{1.2} & 27.1\pmt{0.6} & 41.3\pmt{0.4} \\
 & INSID3 & 49.5\pmt{1.8} & 55.2\pmt{0.9} & 68.0\pmt{1.0} & 55.2\pmt{3.5} & 34.8\pmt{3.2} & 9.4\pmt{0.7} & 23.8\pmt{0.6} & 46.7\pmt{1.5} & 57.0\pmt{1.1} & 28.7\pmt{1.0} & 42.8\pmt{0.3} \\
\rowcolor{oursrow} & \textbf{DINO-KNN} (ours) & 50.1\pmt{1.7} & \textbf{62.2}\pmt{0.7} & \textbf{69.9}\pmt{1.0} & \textbf{60.2}\pmt{2.8} & \textbf{57.2}\pmt{5.5} & \textbf{31.4}\pmt{1.8} & \textbf{61.9}\pmt{0.8} & \textbf{50.7}\pmt{2.0} & \textbf{60.6}\pmt{0.8} & \textbf{50.3}\pmt{0.7} & \textbf{55.5}\pmt{0.2} \\
\midrule
\multirow{3}{*}{15} & GF-SAM & \textbf{51.6}\pmt{2.5} & 54.7\pmt{0.4} & 69.1\pmt{1.2} & 47.9\pmt{2.4} & 34.2\pmt{2.6} & 11.5\pmt{1.6} & 22.9\pmt{0.5} & 40.5\pmt{0.6} & 55.8\pmt{1.0} & 27.3\pmt{1.0} & 41.6\pmt{0.4} \\
 & INSID3 & 49.6\pmt{1.5} & 54.8\pmt{0.8} & 68.9\pmt{0.7} & 55.1\pmt{3.5} & 34.9\pmt{3.3} & 9.2\pmt{0.7} & 23.8\pmt{0.6} & 46.8\pmt{0.9} & 57.1\pmt{0.8} & 28.7\pmt{0.9} & 42.9\pmt{0.4} \\
\rowcolor{oursrow} & \textbf{DINO-KNN} (ours) & 50.4\pmt{1.6} & \textbf{62.5}\pmt{0.6} & \textbf{70.3}\pmt{0.9} & \textbf{61.0}\pmt{2.8} & \textbf{58.9}\pmt{3.2} & \textbf{32.6}\pmt{1.7} & \textbf{62.2}\pmt{0.9} & \textbf{51.2}\pmt{1.8} & \textbf{61.1}\pmt{0.8} & \textbf{51.3}\pmt{0.2} & \textbf{56.2}\pmt{0.3} \\
\bottomrule
\end{tabular}

%% file: sec/4_experiments.tex
\section{Experiments}
\label{sec:experiments}

\subsection{Experimental setup}
\label{sec:protocol}

\paragraph{Datasets.}
We use eight datasets in two groups.
\emph{Natural:} COCO-20$^i$~\cite{Lin:2014:COCO}, LVIS-92$^i$~\cite{gupta2019lvis}, the underwater dataset SUIM~\cite{Islam:2020:Suim} and the aerial dataset iSAID-5$^i$~\cite{Yao:2021:Isaid}.
\emph{Medical:} three ultrasound datasets, CAMUS~\cite{camus} (cardiac ultrasound), FASS~\cite{fass} (fetal abdominal ultrasound) and PSFHS~\cite{chen2024psfhs} (intrapartum transperineal ultrasound), and the surgical dataset EndoVis17~\cite{allan2019endovis} (robotic surgical instruments in endoscopic video).
We report every dataset and three averages: Natural (4), Medical (4) and All (8).

\paragraph{Multiclass episodes.}
In the natural datasets, an episode consists of a target and a reference image that share at least two classes, and the label set is the set of shared classes.
All other annotated classes are treated as background.
In the ultrasound datasets, each image contains all classes, and the references and targets come from different patients.
On EndoVis17, a target frame shows at least two instrument types, and the references come from other videos.
All methods receive the same references for the same target.

\paragraph{Seeds and metrics.}
Every experiment is run with three seeds.
Each seed draws new reference images for the same targets.
We report the mean and standard deviation of mIoU across the seeds. For each dataset, the intersection and union of every class are accumulated over all episodes, and the mIoU is the mean of the per-class IoUs, as in previous work~\cite{cuttano2026insid3}.

\paragraph{Number of episodes.}
We evaluate up to $\nref{=}15$ on every natural target that has fifteen references containing every class of its label set: about 960 on COCO, 250 on LVIS, 1{,}440 on SUIM and 1{,}840 on iSAID.
The medical datasets use 140--150 targets.
The exact numbers per seed are in the supplementary material.

\paragraph{Implementation details.}
\ours{} uses a frozen DINOv3 ViT-L/16 with feature dimension $D{=}1024$.
Images are resized to $1024{\times}1024$ pixels, which gives $P = 64{\times}64 = 4{,}096$ patches per image.
The positional subspace has $d_{\text{pos}}{=}500$ dimensions and the vote uses $k{=}15$ neighbors; \cref{fig:kneighbors} shows that the results change little for $k\ge10$.
These settings are the same for every dataset and every $\nref$.

\paragraph{Baselines.}
We compare with INSID3~\cite{cuttano2026insid3}, GF-SAM~\cite{zhang2024gfsam}, Matcher~\cite{liu2024matcher} and PerSAM~\cite{zhang2024persam}, using the public code of each method.
GF-SAM and Matcher use DINOv2 ViT-L/14 and SAM ViT-H, and PerSAM uses SAM ViT-H, as in their public code.
INSID3 uses the same encoder and input resolution as \ours{}.
These methods predict one binary mask per class.
We run them once per class and merge the masks into a single label map: a pixel claimed by multiple classes is assigned to the class with the highest confidence at that pixel.
PerSAM accepts a single reference.
Matcher is computationally expensive and achieves lower performance. Therefore, we limit its evaluation to $\nref \leq 5$ and report the results in the supplementary material.

\subsection{Main results}
\label{sec:scaling}

\Cref{tab:kscaling} reports the mIoU of every method on each dataset as the number of reference images grows from one to fifteen.
\ours{} has the best average at every $\nref$.

\paragraph{One reference.}
\ours{} has the best average mIoU of $37.8$, followed by INSID3 with $36.1$ and GF-SAM with $34.5$.
INSID3 is better on natural images by $2.9$ points, and \ours{} is better on medical images by $6.4$ points.

\paragraph{More references.}
From $\nref{=}1$ to $\nref{=}15$ the average of \ours{} gains $18.4$ points ($37.8$ to $56.2$), while INSID3 and GF-SAM gain $6.8$ and $7.1$ points and change little after $\nref{=}5$.
On natural images \ours{} is level with INSID3 at $\nref{=}3$ and ahead from $\nref{=}5$ on, with a margin of $4.0$ points at $\nref{=}15$.
On medical images, the margin grows from $6.4$ to $22.6$ points.
From $\nref{=}7$ on, \ours{} is the best method on seven of the eight datasets; the exception is COCO, where GF-SAM leads from $\nref{=}3$ on.

\paragraph{Sensitivity to the references.}
With one reference, the results depend strongly on which reference is drawn: the standard deviation over seeds reaches $9.7$ on CAMUS for \ours{} and $7.5$ for INSID3.
For \ours{} it shrinks from $1.7$ at $\nref{=}1$ to $0.3$ at $\nref{=}15$ on the average.

\begin{figure*}[t]
  \centering
  \includegraphics[width=\linewidth]{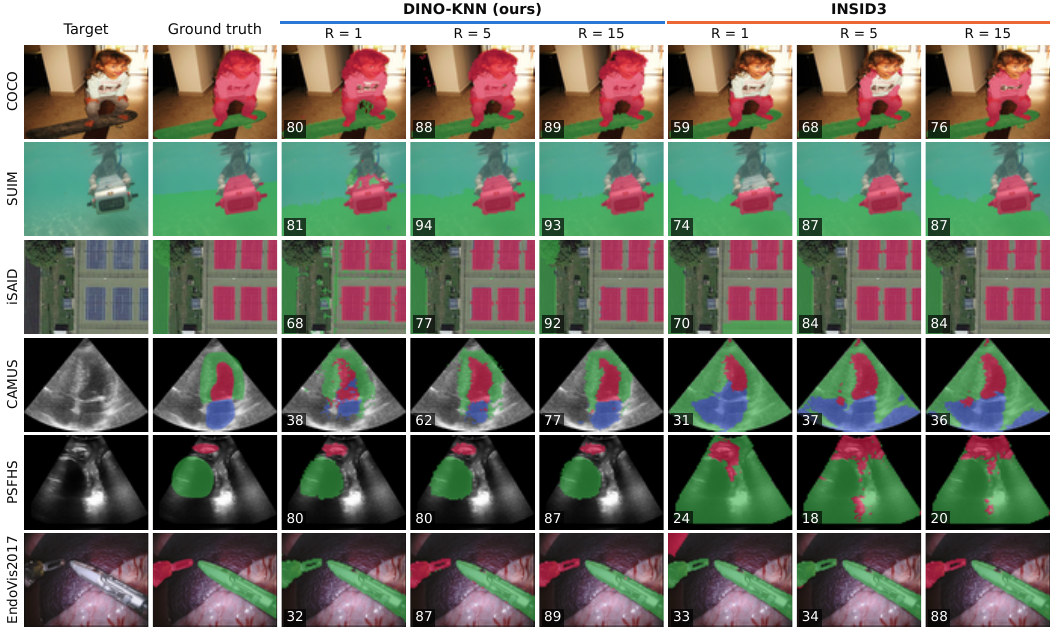}
  \caption{\textbf{Qualitative results of \ours{} and INSID3 for $\nref{=}1,5,15$}, with per-episode mIoU.}
  \label{fig:qual}
\end{figure*}

\paragraph{Qualitative results.}
\Cref{fig:qual} shows selected episodes.
With one reference, the masks of \ours{} are close to the ground truth on COCO, SUIM, and PSFHS, and noisy on CAMUS and EndoVis17.
More references remove most of these errors: on CAMUS the per-episode mIoU rises from $38$ to $77$.

\subsection{Ablation studies}
\label{sec:ablation}

We ablate five design choices of \ours{}: the reference representation, the positional debiasing, the number of neighbors $k$, the encoder size, and the input size.
We also test the setting in which the label set contains every class in the reference, including classes absent from the target.

\begin{table}[t]
  \centering
  \caption{\textbf{Reference representation and positional debiasing.} Average mIoU, mean over three seeds.}
  \label{tab:ablation}
  \footnotesize
  \setlength{\tabcolsep}{1.8pt}
  \resizebox{\linewidth}{!}{\input{tab/ablation}}
\end{table}

\paragraph{Reference representation.}
\Cref{tab:ablation} replaces the $k$-NN vote by two alternatives.
\emph{Nearest prototype} represents each class by the mean of its reference features, and \emph{1-NN} keeps the memory and uses a single neighbor.
The prototype collapses all patches of a class into one vector, so a class whose parts or instances differ in appearance is represented by a mean that matches none of them well.
It reaches $19.0$ with one reference and gains only $2.4$ points with five, because additional references shift the mean but add no new appearances.
1-NN keeps every patch and reaches $35.8$, with a gain of $11.2$ points.
Its decision, however, rests on a single correspondence, so a single wrong match can change the label of a patch.
Voting over $k$ neighbors reduces the effect of such matches and adds $2.0$ points at $\nref{=}1$ and $3.2$ at $\nref{=}5$.

\begin{figure}[t]
  \centering
  \includegraphics[width=\linewidth]{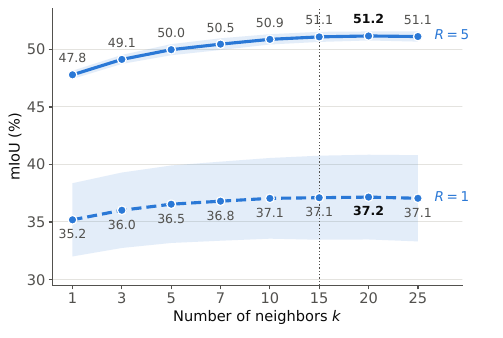}
  \caption{\textbf{Number of neighbors $k$.} Average mIoU of \ours{} over LVIS, CAMUS and EndoVis17, mean over three seeds. Dashed line: $\nref{=}1$; solid line: $\nref{=}5$. The dotted line marks $k{=}15$. Shaded: standard deviation over seeds.}
  \label{fig:kneighbors}
\end{figure}

\begin{figure}[t]
  \centering
  \includegraphics[width=\linewidth]{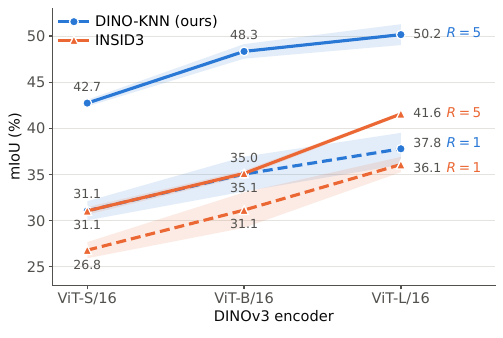}
  \caption{\textbf{Encoder size.} Average mIoU over three seeds with DINOv3 ViT-S/16, ViT-B/16 and ViT-L/16. Dashed lines: $\nref{=}1$; solid lines: $\nref{=}5$. Shaded: standard deviation over seeds.}
  \label{fig:encoder}
\end{figure}

\begin{figure*}[t]
  \centering
  \includegraphics[width=\linewidth]{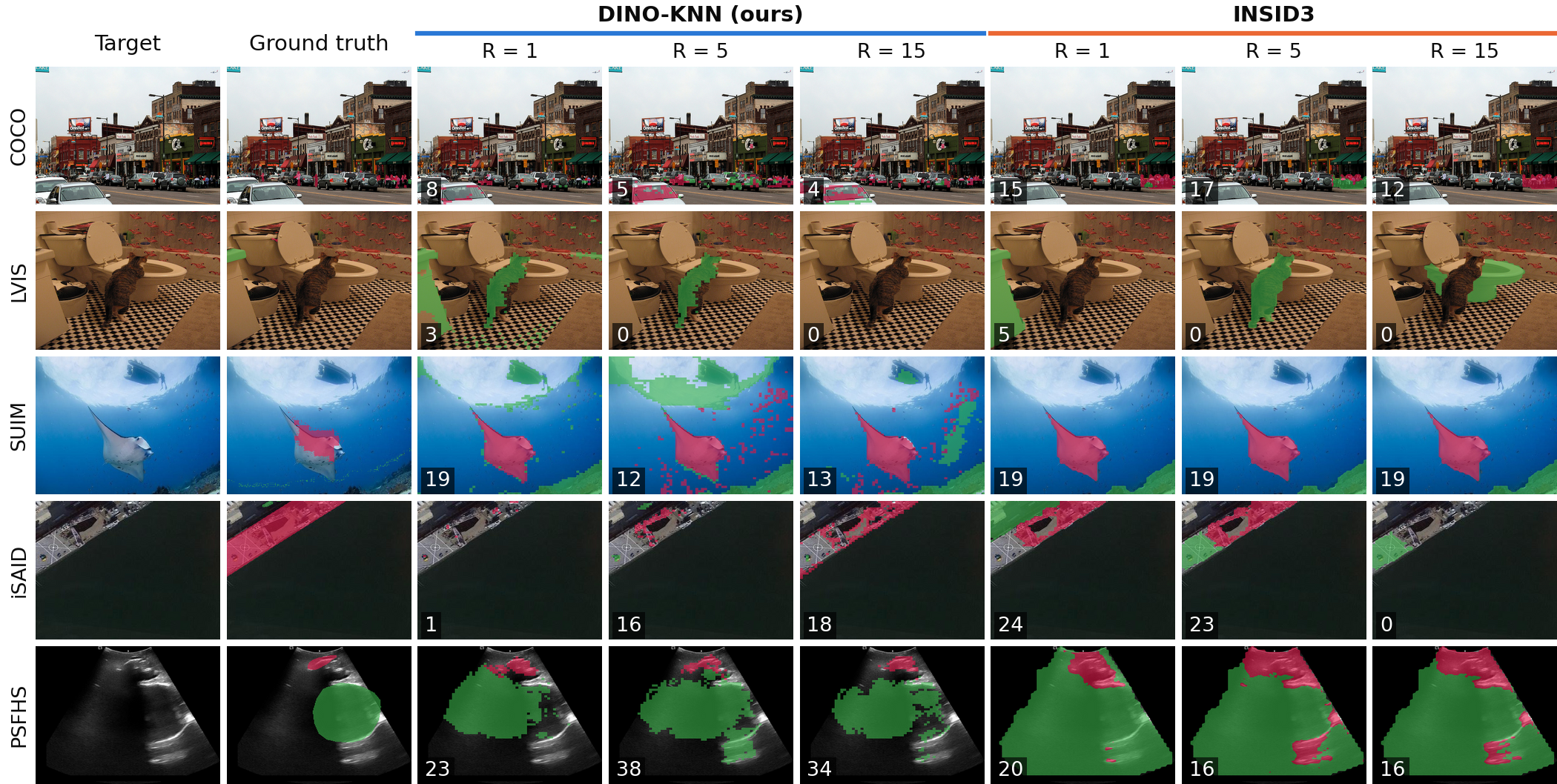}
  \caption{\textbf{Failure cases}: episodes on which \ours{} and INSID3 both have a low IoU at $\nref{=}15$, with per-episode mIoU.}
  \label{fig:fail}
\end{figure*}

\paragraph{Positional debiasing.}
Without the debiasing step (\cref{tab:ablation}) the average drops by $2.6$ points at $\nref{=}1$ and by $2.1$ at $\nref{=}5$.
The loss is about $4$ points on natural images and about $1$ point on medical images, and at $\nref{=}5$ it vanishes on the three ultrasound datasets, as shown in the supplementary material.
We attribute this to the image layout: in ultrasound the anatomical structures appear at similar positions in every image, so the positional component of the features carries useful information, whereas a natural object can appear anywhere.
Removing position is therefore not always beneficial, and for datasets with a fixed layout it can serve as a prior.

\paragraph{Number of neighbors.}
\Cref{fig:kneighbors} varies $k$ on LVIS, CAMUS and EndoVis17.
A single neighbor gives the lowest mIoU ($35.2$ at $\nref{=}1$ and $47.8$ at $\nref{=}5$).
The mIoU rises to $k{=}10$ and changes by at most $0.3$ points beyond it, so \ours{} is not sensitive to the exact choice of $k$.
The variance over seeds is higher at $\nref{=}1$ because the results on CAMUS depend strongly on which reference is drawn; at $\nref{=}5$ it is much lower.

\paragraph{Encoder size.}
We replace ViT-L/16 with ViT-S/16 and ViT-B/16 in \ours{} and INSID3, as shown in \cref{fig:encoder}.
Both methods improve with the encoder size, and \ours{} has the higher average for every encoder and both $\nref$.
With five references, \ours{} with ViT-S ($42.7$) is above INSID3 with ViT-L ($41.6$).
\ours{} is also less sensitive to the encoder size: from ViT-L to ViT-S its average drops by $18\%$ at $\nref{=}1$ and $15\%$ at $\nref{=}5$, against $26\%$ and $25\%$ for INSID3.

\begin{figure}[t]
  \centering
  \includegraphics[width=\linewidth]{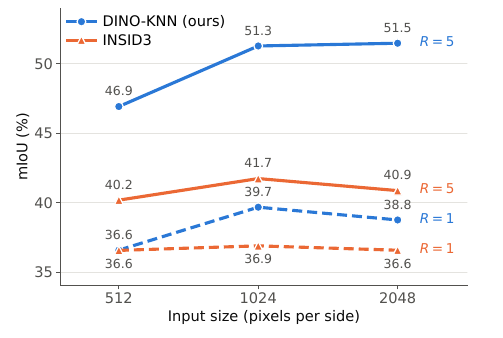}
  \caption{\textbf{Input size.} Average mIoU of \ours{} and INSID3 with $512{\times}512$, $1024{\times}1024$ and $2048{\times}2048$ inputs, one seed, on the targets evaluated at all three sizes. Dashed lines: $\nref{=}1$; solid lines: $\nref{=}5$.}
  \label{fig:inputsize}
\end{figure}

\paragraph{Input size.}
We change the input size of both methods, which changes the patch grid from $32{\times}32$ to $128{\times}128$, as shown in \cref{fig:inputsize}.
From $512$ to $1024$ pixels, \ours{} gains $3.1$ points at $\nref{=}1$ and $4.4$ at $\nref{=}5$, whereas INSID3 gains at most $1.5$ points.
From $1024$ to $2048$ pixels, both methods change by less than $1$ point.
The $1024{\times}1024$ input used in all other experiments is therefore sufficient.

\paragraph{Classes absent from the target.}
So far the label set contains only classes that appear in the target.
In \cref{tab:protocol} the reference keeps all its annotated classes, so some classes of the label set can be absent from the target.
The ultrasound datasets are not used because each image contains every class.
Both methods predict classes that are absent from the target, and every such pixel is a false positive.
Analyzing these predictions on COCO, we find that they replace only background predictions: the mIoU of \ours{} on the present classes is unchanged.
This is because the background is a broad class that covers many different appearances.
Penalizing these false positives lowers the mIoU of both methods by $7$ to $8$ points.

\begin{table}[t]
  \centering
  \caption{\textbf{Classes absent from the target}, mean {\tiny$\pm$std} over three seeds, averaged over COCO, LVIS, SUIM, iSAID and EndoVis17. The references keep all their classes. Predictions of absent classes are either penalized as false positives or ignored.}
  \label{tab:protocol}
  \footnotesize
  \setlength{\tabcolsep}{2.5pt}
  \resizebox{\linewidth}{!}{\input{tab/protocol}}
\end{table}

\begin{table}[t]
  \centering
  \caption{\textbf{Latency per target image and peak GPU memory} on COCO episodes with cached reference features, on one NVIDIA RTX 5000 Ada. Latency is the mean over episodes.}
  \label{tab:runtime}
  \footnotesize
  \setlength{\tabcolsep}{2.6pt}
  \resizebox{\linewidth}{!}{\input{tab/runtime}}
\end{table}

\paragraph{Runtime and memory.}
\Cref{tab:runtime} reports the time to segment one target image, from the loaded image to the final masks, and the peak GPU memory.
All methods run on one NVIDIA RTX 5000 Ada GPU (32\,GB) with $1024{\times}1024$ inputs on COCO.
For a fair comparison with the binary methods, the reference features are computed once and cached for every method.
\ours{} takes $217$ to $225$\,ms, of which about $215$\,ms are the encoder forward pass.
It is $2.1$ times faster than INSID3 with the same encoder and requires $1.5$ to $1.7$\,GB of GPU memory, compared with $5.6$ to $6.8$\,GB for the SAM-based methods.

%% file: tab/ablation.tex
\begin{tabular}{@{}l ccc ccc@{}}
\toprule
 & \multicolumn{3}{c}{$\nref=1$} & \multicolumn{3}{c}{$\nref=5$} \\
\cmidrule(l{2pt}r{2pt}){2-4}\cmidrule(l{2pt}r{2pt}){5-7}
 & Nat. & Med. & All & Nat. & Med. & All \\
\midrule
Nearest prototype & 24.7\pmt{0.4} & 13.3\pmt{2.6} & 19.0\pmt{1.1} & 28.3\pmt{1.0} & 14.5\pmt{2.1} & 21.4\pmt{0.7} \\
1-NN & 41.4\pmt{1.0} & 30.3\pmt{2.5} & 35.8\pmt{1.5} & 52.1\pmt{0.6} & 41.8\pmt{1.8} & 47.0\pmt{1.2} \\
$k$-NN w/o debiasing & 40.3\pmt{1.1} & 30.1\pmt{3.1} & 35.2\pmt{1.7} & 53.2\pmt{0.4} & 43.0\pmt{1.7} & 48.1\pmt{1.1} \\
$k$-NN (ours) & \textbf{44.3}\pmt{1.0} & \textbf{31.3}\pmt{2.8} & \textbf{37.8}\pmt{1.7} & \textbf{56.7}\pmt{0.7} & \textbf{43.7}\pmt{1.6} & \textbf{50.2}\pmt{1.1} \\
\bottomrule
\end{tabular}

%% file: tab/protocol.tex
\begin{tabular}{@{}l cc cc cc@{}}
\toprule
 & \multicolumn{2}{c}{$\nref=1$} & \multicolumn{2}{c}{$\nref=3$} & \multicolumn{2}{c}{$\nref=5$} \\
\cmidrule(lr){2-3}\cmidrule(lr){4-5}\cmidrule(l){6-7}
Method & penalized & ignored & penalized & ignored & penalized & ignored \\
\midrule
INSID3 & \textbf{35.9}\pmt{0.5} & \textbf{44.1}\pmt{1.2} & 42.2\pmt{1.1} & 49.7\pmt{1.2} & 44.2\pmt{1.1} & 51.7\pmt{0.8} \\
\textbf{DINO-KNN} (ours) & 35.2\pmt{0.5} & 42.2\pmt{0.8} & \textbf{43.1}\pmt{0.6} & \textbf{50.7}\pmt{0.7} & \textbf{45.5}\pmt{1.0} & \textbf{53.0}\pmt{0.8} \\
\bottomrule
\end{tabular}

%% file: tab/runtime.tex
\begin{tabular}{@{}l c ccc ccc@{}}
\toprule
 & & \multicolumn{3}{c}{Latency (ms)} & \multicolumn{3}{c}{GPU memory (GB)} \\
\cmidrule(lr){3-5}\cmidrule(l){6-8}
Method & Input & $\nref=1$ & $\nref=3$ & $\nref=5$ & $\nref=1$ & $\nref=3$ & $\nref=5$ \\
\midrule
PerSAM & $1024^2$ & 393 & -- & -- & 5.6 & -- & -- \\
GF-SAM & $1024^2$ & 762 & 488 & 629 & 6.8 & 6.1 & 6.3 \\
INSID3 & $1024^2$ & 470 & 474 & 474 & 1.6 & 1.7 & 1.9 \\
\rowcolor{oursrow}\textbf{DINO-KNN} (ours) & $1024^2$ & 217 & 221 & 225 & 1.5 & 1.6 & 1.7 \\
\bottomrule
\end{tabular}

%% file: sec/5_discussion.tex
\section{Discussion}
\label{sec:discussion}

\ours{} obtains strong results on eight datasets from natural, underwater, aerial, ultrasound, and surgical imaging.
Its main advantage appears when references are added, because every added reference contributes new patches to the memory, whereas a class prototype only shifts its mean (\cref{tab:ablation}).
The other methods saturate after a few references, whereas \ours{} continues to improve up to $\nref{=}15$.
\ours{} is also simple: it uses one frozen encoder and one nearest-neighbor vote, with no second foundation model, no prompt generation, and no mask selection.
These results suggest that combining several foundation models or several processing stages is not necessary for accurate multiclass in-context segmentation.

\paragraph{Limitations.}
With one reference, \ours{} is competitive with INSID3 and GF-SAM on natural images ($44.3$ against $47.2$ and $45.3$ mIoU), and GF-SAM remains better on COCO at every $\nref$.
A class with fewer reference patches than $k$ can be outvoted, which may explain the errors on small objects (\cref{fig:fail}).
Predictions are made per patch, and no boundary refinement is applied.
When the label set contains classes absent from the target, \ours{} still predicts them in background regions (\cref{tab:protocol}).

\paragraph{Failure cases.}
\Cref{fig:fail} shows episodes on which \ours{} and INSID3 both fail with fifteen references.
The most frequent cause is object size: in four of the five rows, one class covers 2 to 18 of the $4{,}096$ target patches and is frequently missed.
On PSFHS, the mask extends beyond the fetal head where its contour is not visible.
In some cases, the ground truth is incomplete, such as in SUIM; \ours{} segments the whole fish, while the annotation covers only part of it.

%% file: sec/6_conclusion.tex
\section{Conclusion}
\label{sec:conclusion}

We introduced \textbf{DINO-KNN}, a training-free and decoder-free approach to multiclass in-context segmentation that labels frozen DINOv3 patch features by a $k$-NN vote over all reference patches. On eight natural and medical datasets, it achieves the best average mIoU for every number of references, and its accuracy keeps growing as references are added: $+18.4$ mIoU from one to fifteen references, against $+6.8$ for INSID3. It is also less sensitive to the encoder size than INSID3. Our ablations show that the gain from more references comes from retaining individual reference patches rather than using a class prototype. These findings establish direct classification of frozen patch features as a strong baseline for training-free multiclass ICS. The patch-level prediction can miss small structures and fine boundaries, motivating future work on spatial detail while preserving the simplicity of the approach.

%% file: sec/7_llm.tex
\section*{LLM Usage Statement}
\label{sec:llm}

A large language model assistant was used in preparing this paper for two purposes: drafting and editing the text, and writing and debugging code.
The method, the experimental design, and the interpretation of the results are the authors' own.
The authors checked the reported numbers against the raw result files and take full responsibility for the content of the paper.

%% file: sec/X_suppl.tex
\section{Comparison including Matcher}
\label{sec:supp_all}

\Cref{tab:supp_all} adds Matcher to the comparison of Tab.~1 of the main paper, for $\nref{=}1,3,5$.
Matcher is evaluated with one seed only because of its computational cost, so all methods in this table use that seed.
PerSAM accepts one reference and appears for $\nref{=}1$ only.
The average mIoU of Matcher is $29.5$, $29.9$ and $29.9$ at $\nref{=}1,3,5$, below GF-SAM, INSID3 and \ours{} at every $\nref$.
\ours{} has the best average at every $\nref$ ($39.7$, $47.2$ and $51.3$).

\begin{table*}[b]
  \centering
  \caption{\textbf{All methods, one seed.} mIoU, $\nref{=}1,3,5$.}
  \label{tab:supp_all}
  \scriptsize
  \setlength{\tabcolsep}{2.6pt}
  \input{tab/supp/s_all_onesed}
\end{table*}

\begin{figure*}[b]
  \centering
  \includegraphics[width=0.92\linewidth]{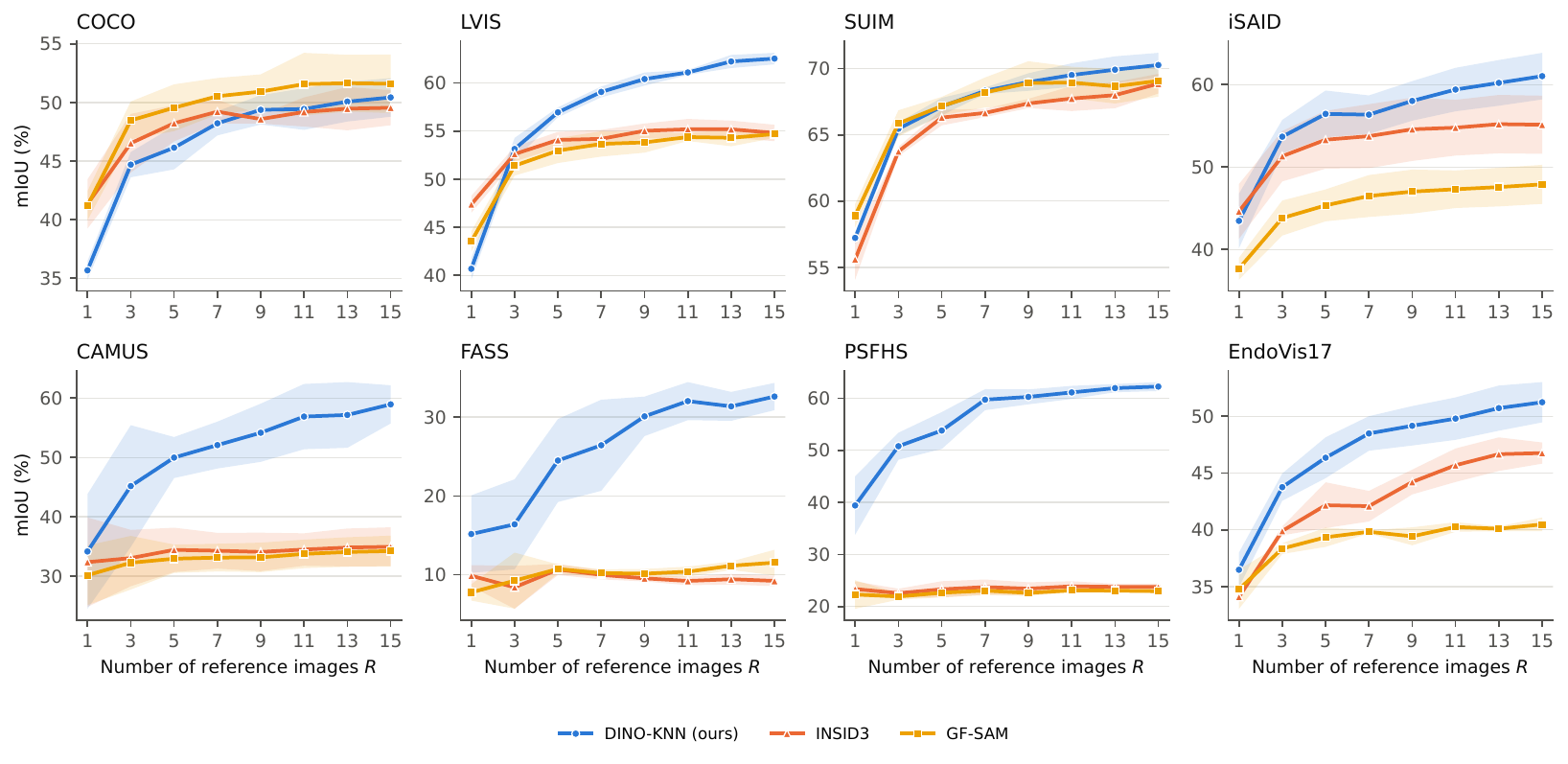}
  \caption{\textbf{mIoU against the number of reference images $\nref$, per dataset} (Tab.~1 of the main paper). Bands: $\pm$ standard deviation over three seeds.}
  \label{fig:supp_scaling}
\end{figure*}

\begin{table*}[t]
  \centering
  \caption{\textbf{Reference representation and positional debiasing in \ours{}}, per dataset (Tab.~2 of the main paper). mIoU, mean {\tiny$\pm$std} over three seeds.}
  \label{tab:supp_refrep}
  \scriptsize
  \setlength{\tabcolsep}{2.6pt}
  \input{tab/supp/s_refrep}
\end{table*}

\begin{table*}[t]
  \centering
  \caption{\textbf{Encoder size} (DINOv3), per dataset (Fig.~5 of the main paper). mIoU, mean {\tiny$\pm$std} over three seeds.}
  \label{tab:supp_encoder}
  \scriptsize
  \setlength{\tabcolsep}{2.6pt}
  \input{tab/supp/s_encoder}
\end{table*}

\begin{table*}[t]
  \centering
  \caption{\textbf{Input size} ($512$, $1024$ or $2048$ pixels per side), per dataset (Fig.~7 of the main paper). mIoU, one seed.}
  \label{tab:supp_input}
  \scriptsize
  \setlength{\tabcolsep}{2.6pt}
  \input{tab/supp/s_inputsize}
\end{table*}

\begin{table*}[t]
  \centering
  \caption{\textbf{Classes absent from the target}, per dataset (Tab.~3 of the main paper). mIoU, mean {\tiny$\pm$std} over three seeds. Predictions of absent classes are either penalized as false positives or ignored. LVIS has no episode with an absent class here, so both metrics are equal.}
  \label{tab:supp_absent}
  \scriptsize
  \setlength{\tabcolsep}{3pt}
  \input{tab/supp/s_protocol}
\end{table*}

\clearpage

\begin{figure*}[p]
  \centering
  \begin{minipage}[t]{0.44\linewidth}
    \centering
    \captionof{table}{\textbf{Number of targets} per dataset and seed in the main comparison.}\label{tab:supp_episodes}
    \footnotesize
    \input{tab/supp/s_episodes}
  \end{minipage}\hfill
  \begin{minipage}[t]{0.52\linewidth}
    \centering
    \captionof{table}{\textbf{Number of neighbors $k$}, per dataset (Fig.~4 of the main paper). mIoU of \ours{}, mean {\tiny$\pm$std} over three seeds.}\label{tab:supp_k}
    \scriptsize
      \setlength{\tabcolsep}{1.6pt}
      \resizebox{\linewidth}{!}{\input{tab/supp/s_kneighbors}}
  \end{minipage}
  \vspace{2em}

  \includegraphics[width=\linewidth]{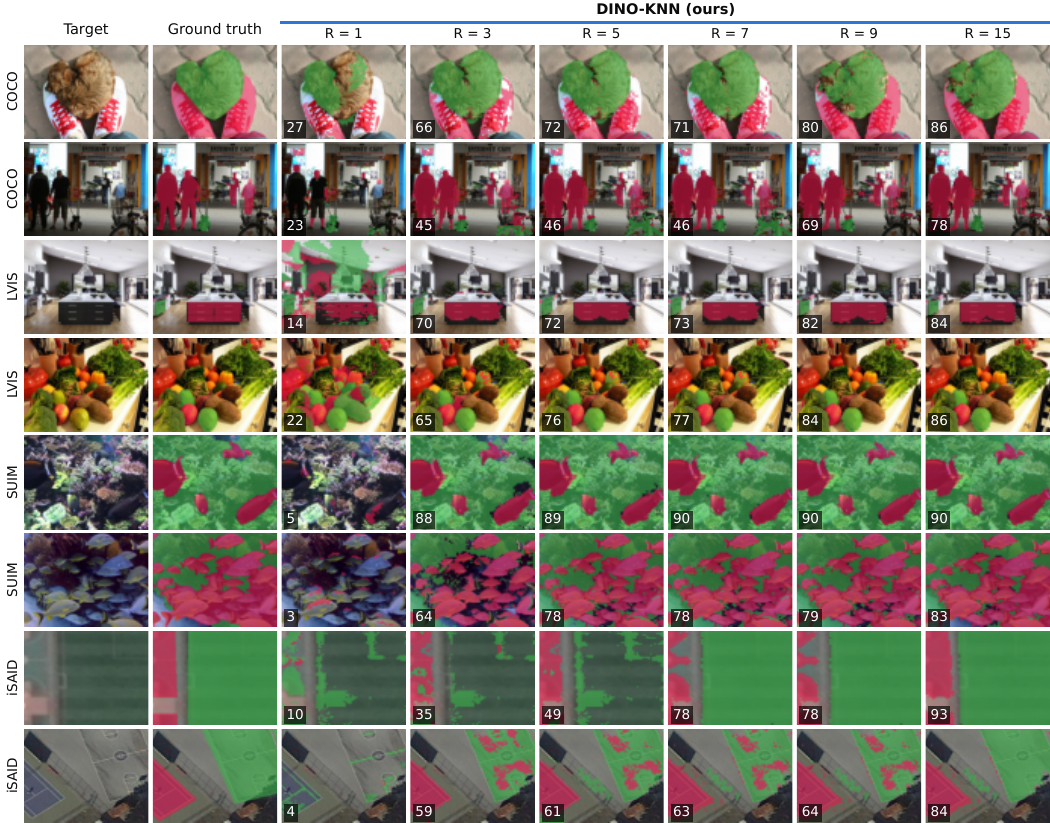}
  \caption{\textbf{Improvement of \ours{} with more references, natural images.}}
  \label{fig:supp_qual_best_nat}
\end{figure*}

\begin{figure*}[p]
  \centering
  \includegraphics[width=\linewidth]{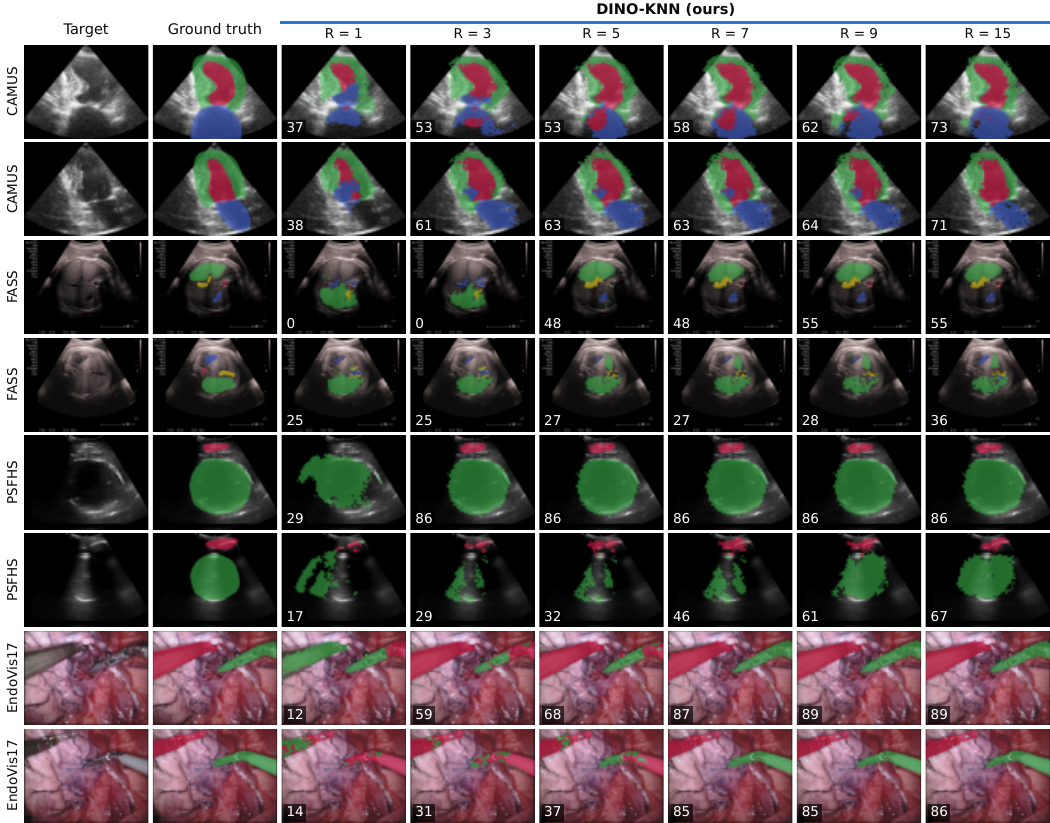}
  \caption{\textbf{Improvement of \ours{} with more references, medical images.}}
  \label{fig:supp_qual_best_med}
\end{figure*}

\begin{figure*}[p]
  \centering
  \includegraphics[width=\linewidth]{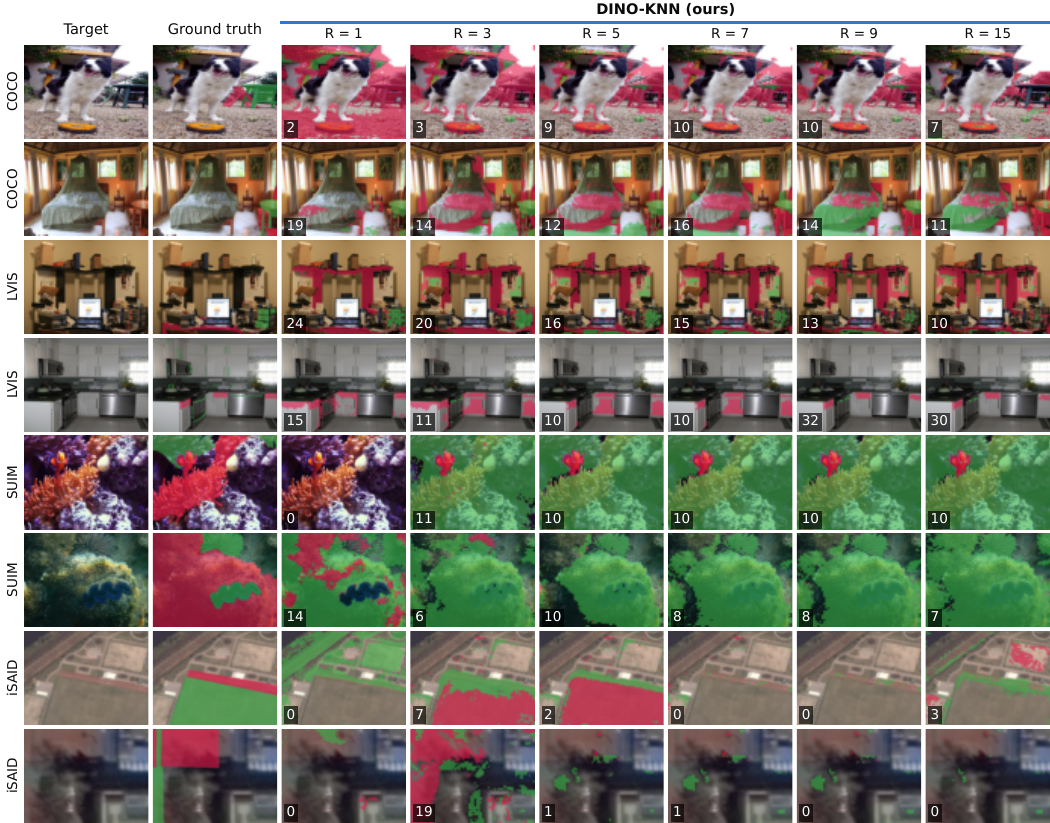}
  \caption{\textbf{Failure cases of \ours{}, natural images.}}
  \label{fig:supp_qual_fail_nat}
\end{figure*}

\begin{figure*}[p]
  \centering
  \includegraphics[width=\linewidth]{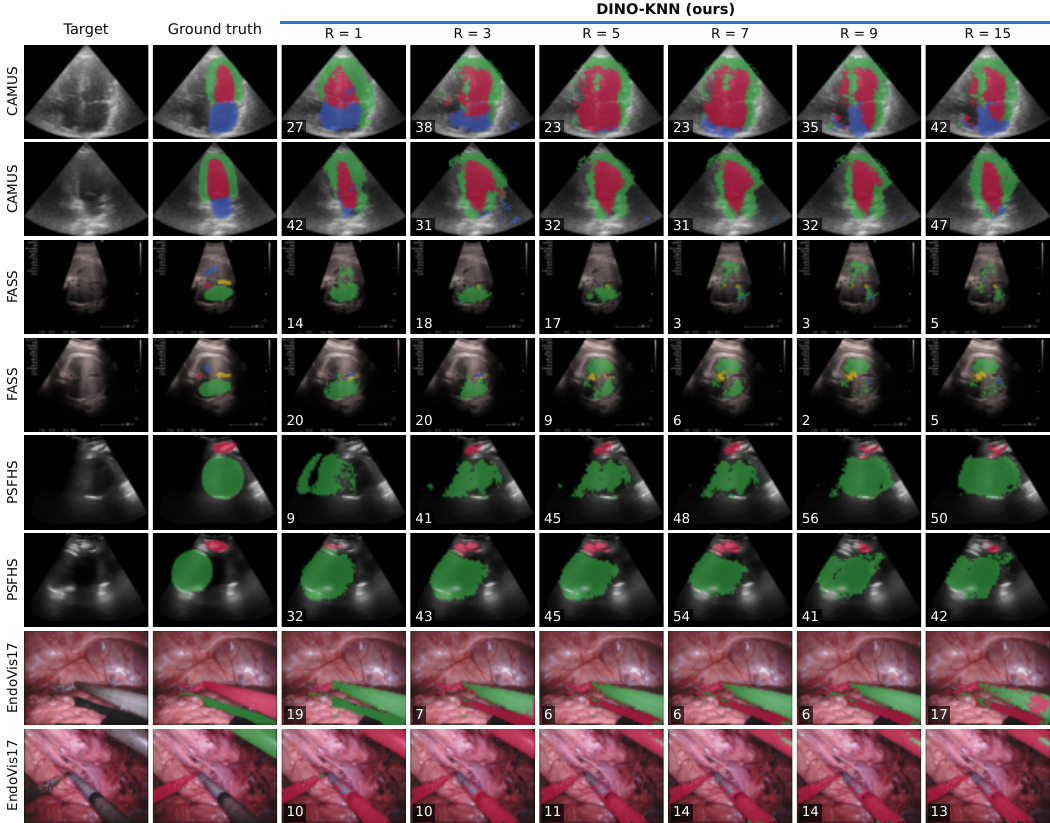}
  \caption{\textbf{Failure cases of \ours{}, medical images.}}
  \label{fig:supp_qual_fail_med}
\end{figure*}

%% file: tab/supp/s_all_onesed.tex
\begin{tabular}{@{}cl cccc cccc ccc@{}}
\toprule
 & & \multicolumn{4}{c}{Natural images} & \multicolumn{4}{c}{Medical images} & \multicolumn{3}{c}{Average} \\
\cmidrule(lr){3-6}\cmidrule(lr){7-10}\cmidrule(l){11-13}
$\nref$ & Method & COCO & LVIS & SUIM & iSAID & CAMUS & FASS & PSFHS & EndoVis17 & Natural (4) & Medical (4) & All (8) \\
\midrule
\multirow{5}{*}{1} & PerSAM & 9.1 & 7.9 & 22.0 & 22.7 & 39.7 & 2.2 & 22.6 & 16.6 & 15.4 & 20.3 & 17.9 \\
 & Matcher & 28.2 & 31.6 & 50.5 & 33.8 & 38.8 & 6.0 & 23.8 & 23.2 & 36.0 & 22.9 & 29.5 \\
 & GF-SAM & \textbf{39.9} & 44.7 & \textbf{58.1} & 38.2 & 34.0 & 7.4 & 24.1 & 36.7 & 45.2 & 25.6 & 35.4 \\
 & INSID3 & 39.1 & \textbf{47.2} & 54.1 & \textbf{47.6} & 39.8 & 8.3 & 24.8 & 34.5 & \textbf{47.0} & 26.8 & 36.9 \\
\rowcolor{oursrow}  & \textbf{DINO-KNN} (ours) & 35.2 & 41.7 & 57.6 & 46.6 & \textbf{43.7} & \textbf{12.8} & \textbf{42.3} & \textbf{37.9} & 45.3 & \textbf{34.2} & \textbf{39.7} \\
\midrule
\multirow{4}{*}{3} & Matcher & 31.3 & 32.0 & 56.5 & 31.9 & 37.1 & 4.7 & 23.0 & 23.2 & 37.9 & 22.0 & 29.9 \\
 & GF-SAM & \textbf{46.7} & 51.1 & \textbf{67.0} & 44.5 & 36.1 & 5.3 & 22.6 & 38.9 & 52.3 & 25.7 & 39.0 \\
 & INSID3 & 44.0 & 53.4 & 63.5 & 53.7 & 37.1 & 5.9 & 23.5 & 39.7 & 53.7 & 26.6 & 40.1 \\
\rowcolor{oursrow}  & \textbf{DINO-KNN} (ours) & 44.0 & \textbf{54.4} & 65.9 & \textbf{55.7} & \textbf{50.5} & \textbf{9.8} & \textbf{53.8} & \textbf{43.7} & \textbf{55.0} & \textbf{39.4} & \textbf{47.2} \\
\midrule
\multirow{4}{*}{5} & Matcher & 30.6 & 31.8 & 54.4 & 30.9 & 34.4 & 9.5 & 22.5 & 25.1 & 36.9 & 22.9 & 29.9 \\
 & GF-SAM & \textbf{47.4} & 52.8 & \textbf{67.9} & 45.4 & 33.9 & 11.4 & 22.4 & 38.4 & 53.4 & 26.5 & 40.0 \\
 & INSID3 & 46.4 & 55.0 & 66.8 & 56.1 & 34.4 & 10.5 & 23.4 & 41.3 & 56.1 & 27.4 & 41.8 \\
\rowcolor{oursrow}  & \textbf{DINO-KNN} (ours) & 44.0 & \textbf{57.4} & \textbf{67.9} & \textbf{59.3} & \textbf{49.7} & \textbf{27.8} & \textbf{57.9} & \textbf{46.0} & \textbf{57.2} & \textbf{45.4} & \textbf{51.3} \\
\bottomrule
\end{tabular}

%% file: tab/supp/s_refrep.tex
\begin{tabular}{@{}cl cccc cccc ccc@{}}
\toprule
 & & \multicolumn{4}{c}{Natural images} & \multicolumn{4}{c}{Medical images} & \multicolumn{3}{c}{Average} \\
\cmidrule(lr){3-6}\cmidrule(lr){7-10}\cmidrule(l){11-13}
$\nref$ & Method & COCO & LVIS & SUIM & iSAID & CAMUS & FASS & PSFHS & EndoVis17 & Natural (4) & Medical (4) & All (8) \\
\midrule
\multirow{4}{*}{1} & Nearest prototype & 17.9\pmt{1.1} & 13.6\pmt{0.2} & 43.3\pmt{1.7} & 23.9\pmt{1.1} & 24.1\pmt{7.0} & 3.6\pmt{1.1} & 14.7\pmt{4.2} & 10.9\pmt{0.9} & 24.7\pmt{0.4} & 13.3\pmt{2.6} & 19.0\pmt{1.1} \\
 & 1-NN & 32.4\pmt{0.4} & 38.2\pmt{1.5} & 53.3\pmt{0.6} & 41.5\pmt{3.4} & 33.1\pmt{9.5} & \textbf{15.9}\pmt{4.5} & 38.1\pmt{4.9} & 34.3\pmt{1.1} & 41.4\pmt{1.0} & 30.3\pmt{2.5} & 35.8\pmt{1.5} \\
 & $k$-NN w/o debiasing & 33.1\pmt{0.2} & 37.3\pmt{0.5} & 52.7\pmt{0.7} & 38.1\pmt{3.9} & \textbf{34.4}\pmt{10.3} & 13.5\pmt{5.1} & 36.7\pmt{7.3} & 35.8\pmt{1.0} & 40.3\pmt{1.1} & 30.1\pmt{3.1} & 35.2\pmt{1.7} \\
\rowcolor{oursrow} & $k$-NN (ours) & \textbf{35.7}\pmt{0.8} & \textbf{40.7}\pmt{1.1} & \textbf{57.2}\pmt{0.4} & \textbf{43.5}\pmt{3.3} & 34.1\pmt{9.7} & 15.2\pmt{4.9} & \textbf{39.4}\pmt{5.6} & \textbf{36.5}\pmt{1.5} & \textbf{44.3}\pmt{1.0} & \textbf{31.3}\pmt{2.8} & \textbf{37.8}\pmt{1.7} \\
\midrule
\multirow{4}{*}{5} & Nearest prototype & 21.2\pmt{1.1} & 14.9\pmt{0.4} & 50.7\pmt{1.7} & 26.4\pmt{1.2} & 25.1\pmt{5.2} & 5.8\pmt{0.8} & 13.7\pmt{3.0} & 13.5\pmt{0.9} & 28.3\pmt{1.0} & 14.5\pmt{2.1} & 21.4\pmt{0.7} \\
 & 1-NN & 41.2\pmt{1.4} & 51.1\pmt{0.5} & 62.5\pmt{0.5} & 53.7\pmt{2.4} & 48.2\pmt{3.0} & 23.4\pmt{6.1} & 51.7\pmt{3.3} & 44.0\pmt{1.7} & 52.1\pmt{0.6} & 41.8\pmt{1.8} & 47.0\pmt{1.2} \\
 & $k$-NN w/o debiasing & 44.3\pmt{1.8} & 52.9\pmt{0.3} & 63.9\pmt{0.6} & 51.8\pmt{2.5} & \textbf{50.3}\pmt{2.4} & \textbf{24.7}\pmt{5.4} & \textbf{53.8}\pmt{4.5} & 43.3\pmt{2.1} & 53.2\pmt{0.4} & 43.0\pmt{1.7} & 48.1\pmt{1.1} \\
\rowcolor{oursrow} & $k$-NN (ours) & \textbf{46.1}\pmt{1.8} & \textbf{56.9}\pmt{0.5} & \textbf{67.1}\pmt{0.7} & \textbf{56.4}\pmt{2.8} & 50.0\pmt{3.5} & 24.5\pmt{5.3} & \textbf{53.8}\pmt{3.5} & \textbf{46.4}\pmt{1.8} & \textbf{56.7}\pmt{0.7} & \textbf{43.7}\pmt{1.6} & \textbf{50.2}\pmt{1.1} \\
\bottomrule
\end{tabular}

%% file: tab/supp/s_encoder.tex
\begin{tabular}{@{}cl cccc cccc ccc@{}}
\toprule
 & & \multicolumn{4}{c}{Natural images} & \multicolumn{4}{c}{Medical images} & \multicolumn{3}{c}{Average} \\
\cmidrule(lr){3-6}\cmidrule(lr){7-10}\cmidrule(l){11-13}
$\nref$ & Method & COCO & LVIS & SUIM & iSAID & CAMUS & FASS & PSFHS & EndoVis17 & Natural (4) & Medical (4) & All (8) \\
\midrule
\multirow{6}{*}{1} & INSID3, ViT-S/16 & 29.3\pmt{0.9} & 32.1\pmt{0.4} & 49.0\pmt{0.5} & 22.5\pmt{1.4} & 32.3\pmt{3.4} & 5.4\pmt{1.8} & 21.3\pmt{5.8} & 22.5\pmt{0.7} & 33.2\pmt{0.6} & 20.4\pmt{2.3} & 26.8\pmt{0.9} \\
 & INSID3, ViT-B/16 & 34.2\pmt{1.6} & 38.0\pmt{0.6} & 56.1\pmt{0.5} & 30.1\pmt{3.3} & 31.7\pmt{8.1} & 7.4\pmt{1.2} & 19.8\pmt{4.1} & 31.8\pmt{2.6} & 39.6\pmt{1.0} & 22.7\pmt{3.4} & 31.1\pmt{1.9} \\
 & INSID3, ViT-L/16 & \textbf{41.3}\pmt{2.1} & \textbf{47.4}\pmt{0.9} & 55.6\pmt{1.5} & \textbf{44.6}\pmt{3.3} & 32.3\pmt{7.5} & 9.9\pmt{1.4} & 23.4\pmt{1.5} & 34.1\pmt{0.3} & \textbf{47.2}\pmt{0.4} & 24.9\pmt{2.0} & 36.1\pmt{0.8} \\
 & DINO-KNN, ViT-S/16 & 28.2\pmt{0.8} & 30.9\pmt{1.2} & 48.0\pmt{0.9} & 27.0\pmt{1.3} & 34.5\pmt{9.5} & 12.5\pmt{4.4} & 35.2\pmt{2.3} & 32.2\pmt{0.2} & 33.5\pmt{0.4} & 28.6\pmt{2.1} & 31.1\pmt{1.0} \\
 & DINO-KNN, ViT-B/16 & 33.5\pmt{0.9} & 36.3\pmt{1.0} & 54.6\pmt{0.7} & 34.0\pmt{1.6} & \textbf{36.2}\pmt{12.1} & 14.4\pmt{4.8} & 34.9\pmt{7.6} & 36.4\pmt{0.9} & 39.6\pmt{0.6} & 30.5\pmt{3.7} & 35.0\pmt{1.9} \\
 & DINO-KNN, ViT-L/16 & 35.7\pmt{0.8} & 40.7\pmt{1.1} & \textbf{57.2}\pmt{0.4} & 43.5\pmt{3.3} & 34.1\pmt{9.7} & \textbf{15.2}\pmt{4.9} & \textbf{39.4}\pmt{5.6} & \textbf{36.5}\pmt{1.5} & 44.3\pmt{1.0} & \textbf{31.3}\pmt{2.8} & \textbf{37.8}\pmt{1.7} \\
\midrule
\multirow{6}{*}{5} & INSID3, ViT-S/16 & 35.6\pmt{1.2} & 37.7\pmt{2.1} & 57.2\pmt{0.5} & 26.2\pmt{1.6} & 34.7\pmt{2.4} & 4.6\pmt{0.2} & 21.8\pmt{1.1} & 30.5\pmt{1.6} & 39.2\pmt{0.5} & 22.9\pmt{0.5} & 31.1\pmt{0.4} \\
 & INSID3, ViT-B/16 & 40.1\pmt{1.8} & 43.1\pmt{0.4} & 64.4\pmt{0.5} & 33.3\pmt{1.6} & 35.1\pmt{2.4} & 6.1\pmt{0.3} & 19.1\pmt{1.2} & 39.6\pmt{3.5} & 45.2\pmt{0.6} & 25.0\pmt{0.1} & 35.1\pmt{0.3} \\
 & INSID3, ViT-L/16 & \textbf{48.2}\pmt{1.5} & 54.1\pmt{0.8} & 66.3\pmt{0.6} & 53.3\pmt{3.5} & 34.4\pmt{3.7} & 10.7\pmt{0.7} & 23.3\pmt{1.6} & 42.2\pmt{2.0} & 55.5\pmt{0.8} & 27.6\pmt{0.7} & 41.6\pmt{0.2} \\
 & DINO-KNN, ViT-S/16 & 37.6\pmt{3.8} & 40.7\pmt{2.1} & 60.8\pmt{0.4} & 41.2\pmt{3.1} & 49.6\pmt{0.8} & 17.4\pmt{3.0} & \textbf{53.9}\pmt{5.9} & 40.6\pmt{2.9} & 45.1\pmt{0.6} & 40.4\pmt{0.3} & 42.7\pmt{0.4} \\
 & DINO-KNN, ViT-B/16 & 44.6\pmt{3.9} & 51.0\pmt{1.6} & 66.1\pmt{1.0} & 46.9\pmt{3.0} & \textbf{55.9}\pmt{0.8} & 21.0\pmt{2.7} & 53.6\pmt{3.9} & \textbf{47.6}\pmt{3.3} & 52.2\pmt{0.7} & \textbf{44.5}\pmt{0.9} & 48.3\pmt{0.8} \\
 & DINO-KNN, ViT-L/16 & 46.1\pmt{1.8} & \textbf{56.9}\pmt{0.5} & \textbf{67.1}\pmt{0.7} & \textbf{56.4}\pmt{2.8} & 50.0\pmt{3.5} & \textbf{24.5}\pmt{5.3} & 53.8\pmt{3.5} & 46.4\pmt{1.8} & \textbf{56.7}\pmt{0.7} & 43.7\pmt{1.6} & \textbf{50.2}\pmt{1.1} \\
\bottomrule
\end{tabular}

%% file: tab/supp/s_inputsize.tex
\begin{tabular}{@{}cl cccc cccc ccc@{}}
\toprule
 & & \multicolumn{4}{c}{Natural images} & \multicolumn{4}{c}{Medical images} & \multicolumn{3}{c}{Average} \\
\cmidrule(lr){3-6}\cmidrule(lr){7-10}\cmidrule(l){11-13}
$\nref$ & Method & COCO & LVIS & SUIM & iSAID & CAMUS & FASS & PSFHS & EndoVis17 & Natural (4) & Medical (4) & All (8) \\
\midrule
\multirow{6}{*}{1} & INSID3, 512 & 36.0 & 44.2 & 54.1 & 47.0 & 38.3 & 9.1 & 28.5 & 35.4 & 45.3 & 27.8 & 36.6 \\
 & INSID3, 1024 & \textbf{39.1} & 47.2 & 54.1 & \textbf{47.6} & 39.7 & 8.3 & 24.8 & 34.5 & \textbf{47.0} & 26.8 & 36.9 \\
 & INSID3, 2048 & 38.8 & \textbf{48.3} & 57.8 & 43.0 & 38.7 & 7.1 & 27.0 & 31.9 & \textbf{47.0} & 26.2 & 36.6 \\
 & DINO-KNN, 512 & 30.8 & 36.2 & 53.8 & 42.7 & \textbf{43.5} & 5.9 & \textbf{44.6} & 35.2 & 40.9 & 32.3 & 36.6 \\
 & DINO-KNN, 1024 & 35.2 & 41.7 & 57.6 & 46.6 & 43.3 & \textbf{12.8} & 42.3 & \textbf{37.9} & 45.3 & \textbf{34.1} & \textbf{39.7} \\
 & DINO-KNN, 2048 & 35.4 & 43.1 & \textbf{58.3} & 44.3 & 43.2 & 10.5 & 40.6 & 34.7 & 45.3 & 32.3 & 38.8 \\
\midrule
\multirow{6}{*}{5} & INSID3, 512 & 43.8 & 51.4 & 62.9 & 55.6 & 34.5 & 8.0 & 23.6 & 41.6 & 53.4 & 26.9 & 40.2 \\
 & INSID3, 1024 & \textbf{46.4} & 55.0 & 66.8 & 56.1 & 34.3 & 10.5 & 23.4 & 41.3 & 56.1 & 27.4 & 41.7 \\
 & INSID3, 2048 & 45.4 & 56.7 & 68.3 & 53.2 & 34.4 & 9.9 & 21.4 & 37.6 & 55.9 & 25.9 & 40.9 \\
 & DINO-KNN, 512 & 40.5 & 51.8 & 65.7 & 55.3 & 49.0 & 16.2 & 55.2 & 41.6 & 53.3 & 40.5 & 46.9 \\
 & DINO-KNN, 1024 & 44.0 & 57.4 & 67.9 & \textbf{59.3} & 49.7 & 27.8 & 57.9 & \textbf{46.0} & \textbf{57.2} & 45.4 & 51.3 \\
 & DINO-KNN, 2048 & 44.1 & \textbf{58.5} & \textbf{68.5} & 56.5 & \textbf{52.3} & \textbf{28.9} & \textbf{58.1} & 44.8 & 56.9 & \textbf{46.0} & \textbf{51.5} \\
\bottomrule
\end{tabular}

%% file: tab/supp/s_protocol.tex
\begin{tabular}{@{}cl ccccc ccccc@{}}
\toprule
 & & \multicolumn{5}{c}{Absent classes penalized} & \multicolumn{5}{c}{Absent classes ignored} \\
\cmidrule(lr){3-7}\cmidrule(l){8-12}
$\nref$ & Method & COCO & LVIS & SUIM & iSAID & EndoVis17 & COCO & LVIS & SUIM & iSAID & EndoVis17 \\
\midrule
\multirow{2}{*}{1} & INSID3 & \textbf{22.1}\pmt{2.0} & \textbf{40.5}\pmt{1.3} & 50.4\pmt{1.0} & 39.6\pmt{3.3} & 27.0\pmt{0.3} & \textbf{44.0}\pmt{1.7} & \textbf{40.5}\pmt{1.3} & 55.2\pmt{1.8} & 47.0\pmt{4.6} & 33.6\pmt{0.1} \\
 & \textbf{DINO-KNN} (ours) & 19.3\pmt{1.5} & 33.1\pmt{0.1} & \textbf{52.2}\pmt{0.9} & \textbf{42.2}\pmt{4.7} & \textbf{29.2}\pmt{1.3} & 37.2\pmt{2.3} & 33.1\pmt{0.1} & \textbf{57.3}\pmt{0.5} & \textbf{47.1}\pmt{3.8} & \textbf{36.4}\pmt{1.5} \\
\midrule
\multirow{2}{*}{3} & INSID3 & \textbf{29.3}\pmt{1.1} & \textbf{45.9}\pmt{0.7} & 57.5\pmt{1.8} & 47.2\pmt{3.8} & 31.2\pmt{0.7} & \textbf{49.1}\pmt{1.5} & \textbf{45.9}\pmt{0.7} & 61.6\pmt{2.0} & 52.2\pmt{3.0} & 39.5\pmt{0.9} \\
 & \textbf{DINO-KNN} (ours) & 26.9\pmt{0.9} & 44.2\pmt{0.6} & \textbf{59.5}\pmt{1.0} & \textbf{51.0}\pmt{3.0} & \textbf{34.1}\pmt{1.0} & 46.0\pmt{2.2} & 44.2\pmt{0.6} & \textbf{64.3}\pmt{1.1} & \textbf{55.8}\pmt{2.1} & \textbf{42.9}\pmt{1.5} \\
\midrule
\multirow{2}{*}{5} & INSID3 & \textbf{32.1}\pmt{2.0} & 46.7\pmt{0.3} & 60.2\pmt{1.6} & 50.5\pmt{4.3} & 31.7\pmt{2.0} & \textbf{52.0}\pmt{1.5} & 46.7\pmt{0.3} & 63.9\pmt{1.5} & 55.1\pmt{2.5} & 40.5\pmt{2.9} \\
 & \textbf{DINO-KNN} (ours) & 28.2\pmt{1.9} & \textbf{48.2}\pmt{1.0} & \textbf{61.1}\pmt{0.8} & \textbf{54.3}\pmt{2.8} & \textbf{35.7}\pmt{1.2} & 47.1\pmt{2.1} & \textbf{48.2}\pmt{1.0} & \textbf{65.9}\pmt{0.8} & \textbf{58.4}\pmt{1.0} & \textbf{45.2}\pmt{2.1} \\
\bottomrule
\end{tabular}

%% file: tab/supp/s_episodes.tex
\begin{tabular}{@{}lccc@{}}
\toprule
Dataset & Seed 1 & Seed 2 & Seed 3 \\
\midrule
COCO & 962 & 965 & 954 \\
LVIS & 255 & 257 & 250 \\
SUIM & 1{,}434 & 1{,}437 & 1{,}436 \\
iSAID & 1{,}841 & 1{,}840 & 1{,}844 \\
CAMUS & 140 & 140 & 140 \\
FASS & 150 & 150 & 150 \\
PSFHS & 150 & 150 & 150 \\
EndoVis17 & 150 & 150 & 150 \\
\bottomrule
\end{tabular}

%% file: tab/supp/s_kneighbors.tex
\begin{tabular}{@{}c cc cc cc cc@{}}
\toprule
 & \multicolumn{2}{c}{LVIS} & \multicolumn{2}{c}{CAMUS} & \multicolumn{2}{c}{EndoVis17} & \multicolumn{2}{c}{Average} \\
\cmidrule(lr){2-3}\cmidrule(lr){4-5}\cmidrule(lr){6-7}\cmidrule(lr){8-9}
$k$ & $\nref{=}1$ & $\nref{=}5$ & $\nref{=}1$ & $\nref{=}5$ & $\nref{=}1$ & $\nref{=}5$ & $\nref{=}1$ & $\nref{=}5$ \\
\midrule
1 & 38.2\pmt{1.5} & 51.1\pmt{0.5} & 33.1\pmt{9.5} & 48.2\pmt{3.0} & 34.3\pmt{1.1} & 44.0\pmt{1.7} & 35.2\pmt{3.2} & 47.8\pmt{0.4} \\
3 & 39.5\pmt{1.4} & 53.3\pmt{0.5} & 33.3\pmt{9.4} & 48.9\pmt{3.1} & 35.2\pmt{1.2} & 45.2\pmt{1.7} & 36.0\pmt{3.3} & 49.1\pmt{0.4} \\
5 & 40.4\pmt{1.5} & 54.8\pmt{0.3} & 33.5\pmt{9.4} & 49.4\pmt{3.1} & 35.7\pmt{1.3} & 45.7\pmt{1.7} & 36.5\pmt{3.4} & 50.0\pmt{0.5} \\
7 & 40.8\pmt{1.3} & 55.8\pmt{0.2} & 33.6\pmt{9.4} & 49.7\pmt{3.2} & 36.0\pmt{1.4} & 45.9\pmt{1.8} & 36.8\pmt{3.4} & 50.5\pmt{0.5} \\
10 & \textbf{40.9}\pmt{1.3} & 56.4\pmt{0.3} & 33.9\pmt{9.5} & 49.9\pmt{3.3} & 36.4\pmt{1.5} & 46.3\pmt{1.8} & 37.1\pmt{3.5} & 50.9\pmt{0.4} \\
15 & 40.7\pmt{1.1} & 56.9\pmt{0.5} & 34.1\pmt{9.7} & \textbf{50.0}\pmt{3.5} & \textbf{36.5}\pmt{1.5} & \textbf{46.4}\pmt{1.8} & 37.1\pmt{3.7} & 51.1\pmt{0.4} \\
20 & 40.8\pmt{1.0} & \textbf{57.2}\pmt{0.7} & 34.2\pmt{9.7} & 49.9\pmt{3.5} & \textbf{36.5}\pmt{1.5} & \textbf{46.4}\pmt{1.9} & \textbf{37.2}\pmt{3.7} & \textbf{51.2}\pmt{0.4} \\
25 & 40.4\pmt{0.9} & \textbf{57.2}\pmt{0.6} & \textbf{34.4}\pmt{9.7} & 49.7\pmt{3.5} & 36.4\pmt{1.6} & \textbf{46.4}\pmt{1.9} & 37.1\pmt{3.7} & 51.1\pmt{0.4} \\
\bottomrule
\end{tabular}